# Through-Wall Pose Imaging in Real-Time with a Many-to-Many Encoder/Decoder Paradigm

Kevin Meng
ACM Student Member
Plano, Texas
kevinmeng@acm.org

Yu Meng, Ph.D.
IEEE Senior Member
Dallas, Texas
yu.meng.us@ieee.org

***Abstract*— Overcoming the visual barrier and developing "see-through vision" has been one of mankind's long-standing dreams. Unlike visible light, Radio Frequency (RF) signals penetrate opaque obstructions and reflect highly off humans. This paper establishes a deep-learning model that can be trained to reconstruct continuous video of a 15-point human skeleton even through visual occlusion. The training process adopts a student/teacher learning procedure inspired by the Feynman learning technique, in which video frames and RF data are first collected simultaneously using a co-located setup containing an optical camera and an RF antenna array transceiver. Next, the video frames are processed with a computer-vision-based gait analysis "teacher" module to generate ground-truth human skeletons for each frame. Then, the same type of skeleton is predicted from corresponding RF data using a "student" deep-learning model consisting of a Residual Convolutional Neural Network (CNN), Region Proposal Network (RPN), and Recurrent Neural Network with Long-Short Term Memory (LSTM) that 1) extracts spatial features from RF images, 2) detects all people present in a scene, and 3) aggregates information over many time-steps, respectively. The model is shown to both accurately and completely predict the pose of humans behind visual obstruction solely using RF signals. Primary academic contributions include the novel many-to-many imaging methodology, unique integration of RPN and LSTM networks, and original training pipeline.**

## I. INTRODUCTION

Our ability to perceive information about our environment suffers from a significant bottleneck from the physical properties of visible light: it is either reflected or absorbed by objects in our immediate surroundings, allowing various items such as furniture and walls obstruct our view of entities we want to see. Especially in search & rescue, non-invasive healthcare, and military operations, the ability to perceive human presence and recover the figure could prove instrumental to saving lives.

However, most types of electromagnetic radiation other than visible light are either too powerful, to the extent of causing adverse health effects by exposure, or too high in wavelength, causing them to pass straight through objects. In contrast, Radio Frequency (RF) electromagnetic radiation is safe, can traverse materials of low reflective index such as walls and furniture, and reflect off humans. This makes them an ideal imaging wave to harness in transcending the physical limits of traditional vision.

There are many applications for the solution proposed in this paper, one of which is Search & Rescue: RF can detect victims behind an assortment of obstructions, potentially preventing a portion of the 1.35 million deaths and 218 injuries caused by 6,873 natural disasters worldwide between 1994 and 2013 [1].

An especially exciting application of such a technology is in fire search & rescue. In environments polluted by smoke, infrared radiation, and physical debris, no technological instruments can currently image life from safety. RF imaging could provide not only detection but also figure retrieval [11].

This paper aims to establish an innovative methodology that allows people to detect the human figure through visual obstruction. Figure 1 depicts the intended system design. First, RF signals are transmitted and received by an RF imaging device. While some signals are reflected, absorbed, or attenuated by an intermediate obstruction, others penetrate and reflect off a human subject. Using a computing device, RF reflection data is extracted and processed, before being inputted into a deep learning pose decoding module.

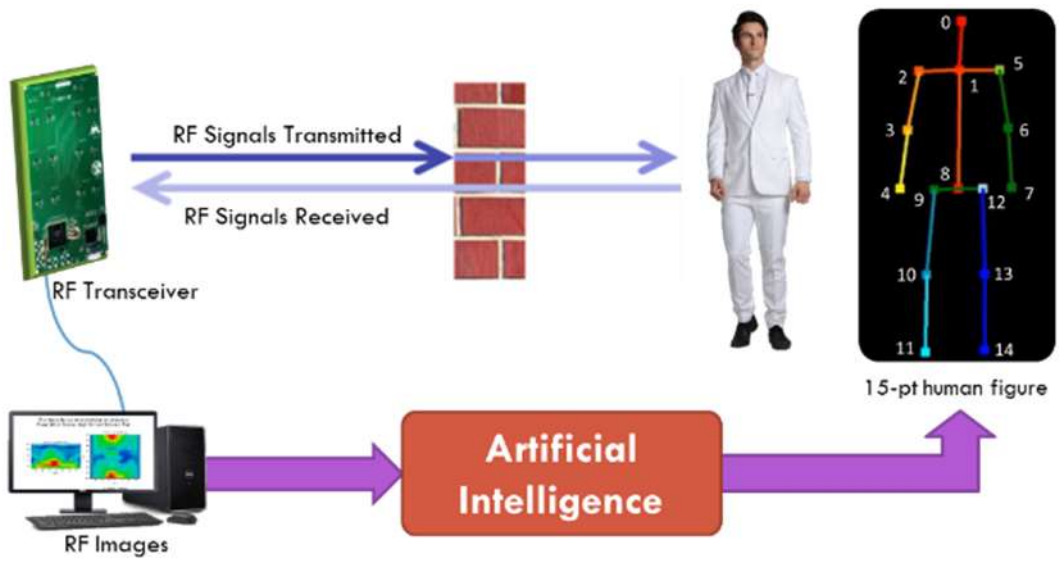

Figure 1. Diagram of intended design

## II. RELATED WORKS

### A. *Figure Retrieval Using Physical Sensors*

Estimation of the human figure via physical sensors has been developed extensively and put into commercial production [37] due to the relative ease with which the data may be extracted using inertial measurement units [38]. However, these solutions require the target user to wear a bevy of cumbersome sensors on their bodies. In some cases, including search & rescue cases or military operation, this bottleneck defeats the purpose of trying to estimate the human figure. In other cases, this causes a major inconvenience for users, as in medical healthcare.

### B. *Figure Retrieval Using the RGB Domain*

Analyzing the human figure using RGB video has also been a hotspot of recent computer vision research, as the high spatial resolution of color images allows for an accurate and complete extraction of the human pose. Notably, Zhe Cao et al. [20] have developed a state-of-the-art pose estimation technique using Part

Affinity Fields to extract 2D poses from images. Another group developing similar systems is Microsoft Asia Research, with their Microsoft Human Pose Estimation (MSFT) model [47].

### *C. Figure Retrieval Using the RF Domain*

In the past, RF signals ( $3\text{kHz} - 300\text{kHz}$ ) were used sparingly in human localization problems, because they required the user to carry a Wi-Fi or Cellular-enabled device to query the person's location [39, 40]. This provided little advantage relative to the solutions mentioned in §II-A. Recently, however, researchers have begun studying the usage of RF signals in an imaging context, as opposed to their traditional function as a communications wave. In these works, RF signals are broadcast into the environment, and their reflections off human bodies are used to deduce the human figure behind walls. These works can be stratified into two primary categories. First is the research concerning high-precision imaging using $100\text{GHz}$ frequencies (colloquially known as mmWave). This provides precise detail but is incapable of penetrating walls or furniture. In contrast, the second category of work concerns penetrative applications using Wi-Fi-band signals ($3\text{GHz} \leq f \leq 10\text{GHz}$). These EM waves provide less detail but allow the penetration of common building materials. Discussions will focus on Wi-Fi-band imaging for its penetrative capability, which is central to this study.

Currently, few reflection-based RF imaging systems can provide high detail; most works can only achieve localization [41, 42, 43] or coarse approximation of the human figure, such as in [7]. Limitations to reflective RF imaging are caused by physical properties of RF waves, including specularity and low spatial resolution. For this reason, poses recovered from models may exhibit incompleteness and flickering. [35] attempted to address specularity using 3D convolutions. However, studies have shown jointly convolutional and recurrent models to capture spatio-temporal dependencies better [36]. Additionally, no previously proposed deep learning system exhibiting through-wall vision capability is trained using a pipeline *explicitly* designed to handle attenuation and noise associated with wave propagation in complex media. In this paper, we present a novel approach that aims to resolve shortcomings of previous works.

## III. PROCEDURAL FLOW

This paper proposes a system in which RF signals are computationally analyzed by a deep learning model to infer the pose of person(s) though visual obstruction, including walls and furniture.

An analysis of the scientific problem yields the formulation of a 5-step project flow: 1) Device Sensors, 2) Data Acquisition, 3) RF Data Pre-Processing, 4) Deep Learning Modeling, and 5) Output Performance Analysis. Each of the following sections will explain the motivation and implementation behind design choices in this paper.

## IV. DEVICE SENSORS

Sensory data is collected simultaneously in two separate channels: RF reflection signals and RGB images. The RF signals will be used as input to the deep learning model during training and inference, while the RGB images will be inputted to a second computer vision model during training to cross-modally supervise the RF image-to-pose model. To ensure that images from both the camera and webcam are representative of the same scene from the same viewing angle, the FMCW antenna array and RGB camera are mounted in positions fixed relative to each other on a tripod. Furthermore, to ensure synchronization in the time domain, the RF and RGB inputs are synced to within ~5ms of error.

### *A. FMCW Radio-Frequency Antenna Array*

This paper utilizes an RF antenna array as the imaging device, which transmits signals on the RF wavelength into the environment and collects reflected signals. The antenna array used in this paper is the Walabot Developer, a commercially-available, FCC-compliant radio. Technical specifications include a frequency bandwidth in range 3.3GHz to 10GHz and transmit power of approximately $100pW$, which is $1/1000$ the transmission power of Wi-Fi. Its API returns a 3D distribution, where each point represents the signal power received at that voxel. Imaging is completed on the spherical coordinate system $(\theta, \phi, r)$. We refer the reader to [2] for more details on the hardware device.

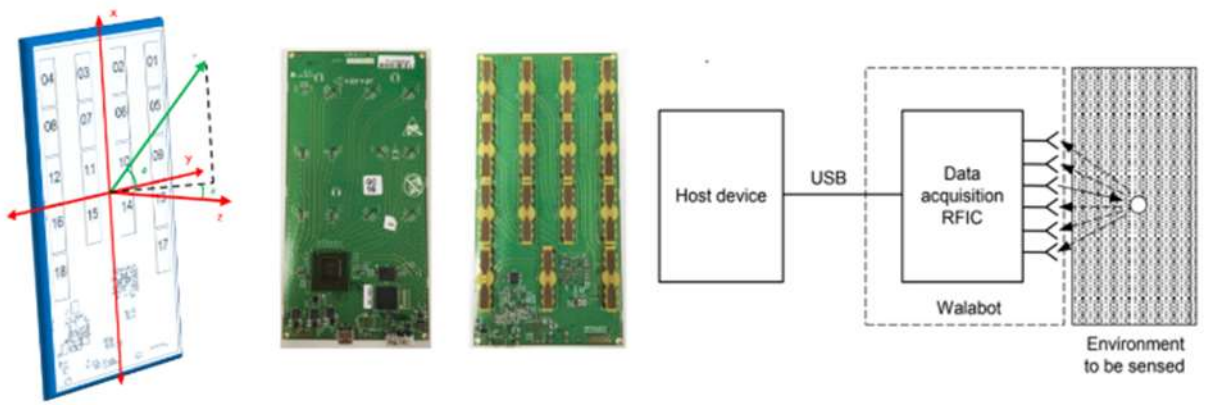

Figure 2. Radio-Frequency Antenna Array

### *B. RGB Camcorder*

The Logitech HD Pro Webcam C920 is used to capture full-color RGB video of the environment. It supports connection to computers via the USB-A protocol. The camera resolution is downgraded to 640x480 to reduce storage consumption.

### *C. Co-Located Setup with Array and Camera*

The Camera and Array are spatially co-located to ensure that the outputs are consistent. Table 1 details the data collection from the various sensory inputs.

Table 1. Sensory Input, Data Acquisition, and Data Format

| Sensors | Data Acquisition | Data Format |
|---|---|---|
| FMCW Antenna Array | RF reflections from environment | .csv file |
| RGB Camera | RGB image of human(s) in environment | .mp4 file |

## V. DATA ACQUISITION VIA RF SENSING

In this study, a primary challenge lies in determining a viable physical means to overcome the boundaries imposed on vision by the ~380nm to ~750nm wavelength range of the electromagnetic spectrum, better known as visible light. At this wavelength, it fails to penetrate opaque objects. Alternatively, radio frequency (RF) waves, such as Wi-Fi signals, *can* penetrate walls. In addition, humans have high reflective coefficients [3]; therefore, RF waves can be used to break the vision barrier. To harness these types of EM waves, we employ antenna arrays.

The following subsections will detail the implementation and key operative features of necessary imaging devices.

### *A. Single Transmitter and Recevier Antenna Pair*

An RF signal is a waveform which has a periodic phase factor. By calling or sampling the received signal, both amplitude and its phase can be recorded [4]. The antennas in the radio used in this study have broadband performance covering the 3-10GHz frequency range.

### *B. Antenna Array*

Antenna arrays can be used to identify the spatial direction from which an RF signal arrives with significantly improved ability to discern spatial direction compared to a single antenna. The angular resolution of an antenna array can be expressed as:

$$\Delta\theta = 0.886\frac{\lambda}{nd} \tag{1}$$

where $n$ is number of transmitters, and $d$ is the space interval between adjacent antennas [7]. We refer readers to [6] for details regarding antenna arrays.

### *C. Frequency-Modulated Continuous Wave (FMCW)*

Although antenna arrays, by themselves, are able to perceive spatial location, they are not sensitive to depth. A straightforward method of measuring depth is using the distance formula: $d = ct$, where $c$ is the speed of light. However, the time of flight $t$ is extremely difficult to measure, as it is typically in the nanosecond range [8].

FMCW is a special radar technique that allows the measurement of reflector depth in a feasible manner. Rather than attempting to capture the miniscule difference between time of transmission and receipt, it sends a signal linearly modulated in frequency w.r.t. time. In this manner, rather than measuring the time delay between the transmitted chirp and reflected chirp, the $\Delta f$ is measured; this can be accomplished using a low-cost, passive hardware device called a mixer [9]. To compute $\Delta t$ given $\Delta f$, we simply use the modulation slope $k$:

$$\Delta t = \frac{\Delta f}{k} \tag{2}$$

Equipped with this knowledge, we can compute the reflected power emanating from a depth $r$. Furthermore, depth resolution depends on the bandwidth of the frequency chirps, as shown in the equation below:

$$\Delta r = \frac{c}{2B} \tag{3}$$

where $B$ is the difference between the maximum and minimum frequencies of the chirps [11].

Jointly, antenna arrays and FMCW allow us to perceive 3D information through RF reflection, analogous to an optical imaging system with visible light reflections.

### *D. Antenna Array SDK API for Implementation*

Following the hardware considerations, we now examine its interface with software. The Walabot API, contained in Walabot Developer Pack SDK [12], is used to grab data from the RF hardware. Data is transposed into a 3D matrix representing reflected signal power, and $R(\theta, \phi, r)$ is used to query the raw reflected signal power from each voxel in 3D space.

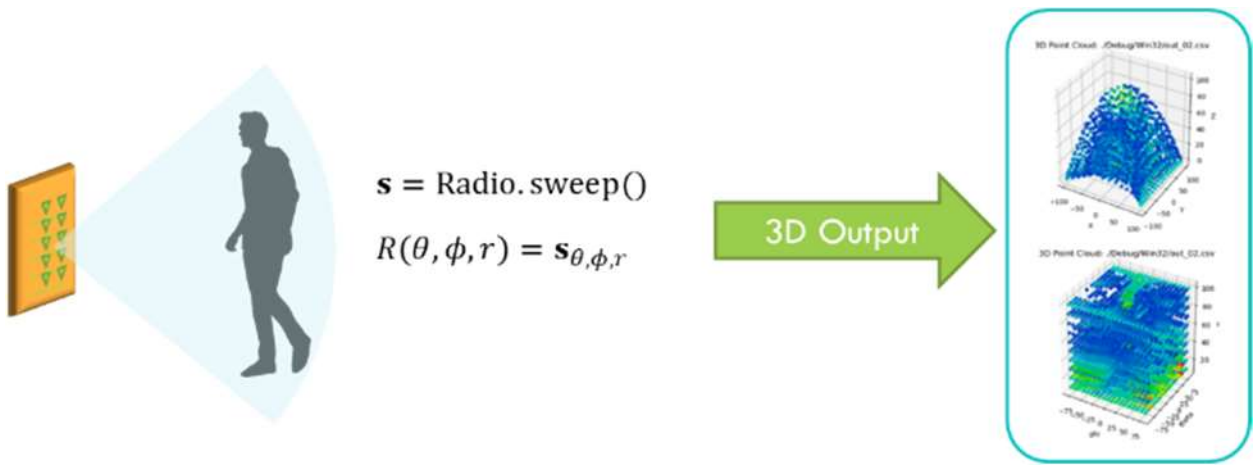

Figure 3. Collection Setup

## VI. RF Data Pre-Processing

### *A. Coordinate Conversion*

Sensor data is retrieved in spherical coordinates using API calls to the Walabot Developer SDK. However, Cartesian representation is desired to represent 3D space [13]. Equation 4 is used to convert between the two coordinate systems.

$$\begin{cases} x = r * \sin\theta \\ y = r * \cos\theta * \sin\phi \\ z = r * \cos\theta * \cos\phi \end{cases} \tag{4}$$

### *B. Dimensionality Reduction*

A 3D point cloud is computationally expensive to compute over and potentially prohibitive to real-time execution. To reduce computational demand, we propose to simplify the 3D point clouds collected by the RF antenna array into two 2D heatmaps: vertical and horizontal. By summing values of reflection intensity over two planes, we minimize the number of sparse data points, while keeping features salient to the decoding of bodily keypoints. Reduction in the computational complexity by an order of magnitude significantly reduces time and space demands.

$$\begin{cases} R_{horz}(x, z) = \sum_{y=y_{min}}^{y_{max}} R(x, y, z) \\ R_{vert}(x, y) = \sum_{z=z_{min}}^{z_{max}} R(x, y, z) \end{cases} \tag{5}$$

## VII. Physics-Driven Design Considerations

In §VI, the final step of RF data processing is performed. From this step, both horizontal and vertical heatmaps are outputted to represent the person in the RF domain. However, analysis of vertical heatmaps, in particular, reveals a significant challenge: not all body parts show up on a single RF heatmap. Within 200ms, body parts with smaller surface areas can disappear.

This section aims to A) identify the physical cause of this phenomenon and B) outline a solution that employs state-of-the-art deep learning techniques.

### *A. Body-Part Specularity (Reflection and Scattering)*

Reflection occurs when a beam impinges upon a surface smooth relative to its wavelength; the physical behavior of the wave will obey the law of reflection, i.e. $\theta_{incident} = \theta_{reflected}$ [14]. On the other hand, scattering occurs when a wave impinges upon an object that is rough or uneven relative to the signal's wavelength, causing the reflected energy to spread out or "scatter" [15]. Figure 4 depicts this phenomenon.

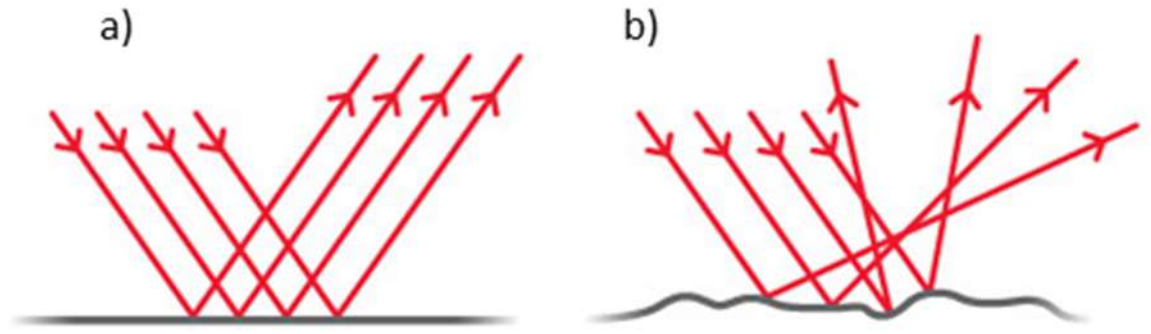

Figure 4. Contrast Between Reflection and Scattering

In experimental settings, the $\lambda$ of RF is ~5cm, as opposed to the ~$10^{-7}$ m wavelength of visible light. Therefore, the physics of RF imaging are fundamentally different from that of optical imaging. With respect to the miniscule wavelength of visible light, surfaces are rough and therefore function as scatterers, as seen in Figure 4(b). Conversely, in the ~5cm wavelength of RF, objects function as reflectors. Therefore, only signals that fall approximately normal to an object surface are reflected back toward the transceiver. However, as the person walks, different body parts reflect signals toward the device and become visible to the device at different timesteps.

### B. *Deep Learning-Based Solution Sketch*

Based on observations made in the previous subsection, we deduce that RF imaging, unlike optical imaging, is *not a one-to-one scenario*. In optical imaging, one sampling of lens-focused light waves is sufficient to reconstruct a full image; this is because objects scatter light waves. However, in the RF domain, not all critical limbs show up at once. Therefore, to compute the human figure at each timestep, the model must consider not just the current RF image, but also several that come before it. This enables the model to together various limbs that reflect signals back to the receiver at various times.

A computational method by which this can be accomplished is deep learning. Namely, Recurrent Neural Networks (RNN) [16] have been the workhorses behind recent breakthrough applications in analyzing sequential data, because they consider past information when constructing each prediction. In this manner, RNN's are capable of accumulating information over time to make an accurate pose inference. RNN's can also be combined with Convolutional Neural Networks (CNN) [17], which learn spatial features from RF heatmaps. In tandem, CNN's and RNN's jointly learn spatio-temporal patterns.

## VIII. Deep Learning Architecture & Algorithms

### A. *Body Pose Keypoint Definitions*

Figure 1 depicts the BODY-15 definition of the 15 body keypoints that are to be detected through visual obstruction. These include: head, shoulder, elbow, hand, hip, knee, and foot datapoints. At each timestep, each of these keypoints will be classified into a pixel on the screen. Note that there exist more granular definitions of body keypoints [20], but the body-part specularity phenomenon and low spatial resolution of RF images do not lend to such detailed representations.

### B. *Overview of Cross-Modal Supervision Algorithm*

Because it is infeasible to manually label radio-frequency image with the appropriate keypoints, a more efficient manner of generating supervisory labels is desired. This involves a computational method utilizing a deep learning model cross-modally dependent on another.

Let $\mathcal{M}_1$ denote an initially untrained deep learning model (i.e. See-Through Model) that takes RF images as input and outputs 15 body keypoints at each time step. Let $\mathcal{M}_2$ denote a pre-trained deep learning model that takes RGB images as input and outputs 15 body keypoints at each time step.

$\mathcal{M}_1$ will be cross-modally supervised by $\mathcal{M}_2$ during training. More concretely, each training sample consists of only an input $x_i$, which contains 5 channels of data: 2 for the horizontal and vertical heatmaps, and 3 for the RGB image. All 5 channels describe the same moment in time from different perspectives. A supervision label $y_i$, containing, the "ground truth" values for each of the 15 keypoints, will be generated with the evaluation of $\mathcal{M}_2(x_i[\text{RGB}])$. Then, the training pair $(x_i[RF], y_i)$ will be inputted into $\mathcal{M}_1$, resulting in final prediction vector $\hat{y}_i$. Loss will be computed as $J(y_i, \hat{y}_i; \theta)$, where $J$ is the objective. In practice, we implement $\mathcal{M}_2$ as the state-of-the-art OpenPose computer vision gait analysis module.

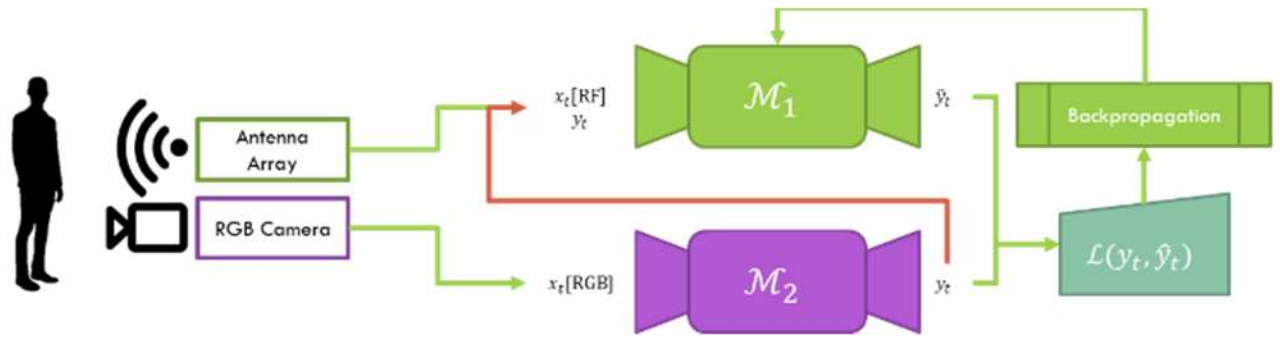

Figure 5. Cross-Modal Supervision

### C. *Novel Training Pipeline with Artificial Obstruction*

A supplemental stand is used to introduce RF obstruction without blocking the RGB camera: Wood, brick, drywall, concrete, plastic, paper board, insulation, linoleum, carpet, fog, leaves. This increases robustness of trained model through explicit learning on through-obstruction scenes.

### D. *Multilayer Perceptron (MLP) & Deep Neural Network*

Neural networks [22] are composed of layers of neurons, much like the biological brain. Each neuron accepts input $p$ and outputs $a = wp + b$. $a$ is then subject to an activation function $\phi(\cdot)$ that regularizes its value and introduces non-linearity; this produces final output $a'$. $w$ and $b$ are values that can be adjusted through training of the network. Neurons can have many inputs, in the form of vectors. If $\mathbf{p}$ is a vector of inputs, $\boldsymbol{\ell}$ is a layer of neurons, $\mathbf{W}$ is a weight matrix representing all connections between $\mathbf{p}$ and $\boldsymbol{\ell}$, and $\mathbf{b}$ is a vector of bias values for each output, then the operations of a neuron are best represented by matrix multiplication; output vector $\mathbf{a}' = \phi(\mathbf{W}\mathbf{p} + \mathbf{b})$.

### E. *Residual Convolutional Neural Network*

In contrast to deep multilayer perceptron models where each layer is fully connected to the layers around it, each neuron in a convolutional neural network (CNN) [17] is only locally connected with a few neurons in the surrounding layers. Additionally, all neurons in a CNN layer share identical weights. Mathematically, the value of a neuron in a CNN can be computed as the convolution of a weight kernel with the neurons in the previous layer:

$$a_n = \phi(k_n * a_{n-1}) \quad (6)$$

where $*$ is the convolution operator, and $k_n$ is the weight kernel at layer $n$. CNNs leverage local dependencies and features in data, especially that of images, to reduce the total number of learned parameters. This acts as a regularization and resource reduction mechanism. The neurons, or weight kernels $k_n \forall n$, in a CNN are referred to as feature maps, as they can be viewed as features that correspond to different parts of the input. Deeper layers, or layers with a large $n$, will learn increasingly abstract and global properties of the image [23].

A derivative of the CNN, the Residual Convolutional Neural Network (ResNet) [24], is employed to enable deeper stacking of layers without high risk of overfitting, all while smoothing the topology of the objective function for a more favorable optimization landscape. ResNet is implemented as DarkNet-53, the convolutional foundation of the YOLOv3 detection algorithm.

### *F. Region Proposal Network for Multiple People per Frame*

A Region Proposal Network is adopted to detect multiple people within a single snapshot, frame-by-frame. The RPN is implemented as in the YOLOv3 algorithm; we refer the reader to [25] for more information on the object detection technique. In short, YOLOv3 detects objects by sliding a convolutional network over the final ResNet output feature map.

The RPN accepts features from the CNN and outputs a $P \times D$ size matrix, where $P$ is the maximum number of people in a frame and $D$ is the dimensionality of data encoding each person in the RF image. In this paper, $D = 445$ for the 4 bounding box descriptors [25], denoted $\mathbf{d}$, and $21 \times 21$ sized flattened feature map, denoted $\boldsymbol{\sigma}$. Variable-sized feature maps are cropped and resized to the designated size using Region of Interest (ROI) Pooling, as in [25]. The 445 values are expressed in a concatenated one-dimensional vector.

Now we define an objective to minimize for the RPN, for which two factors are considered: 1) localization error: the discrepancy between 4 corners of the bounding box should be minimized, and 2) continuity error: predicted bounding boxes should not dart across space, so a small portion of the loss is dedicated to minimizing the L1 distances between bounding box vertices. To improve convergence, continuity error is modified by a factor of 0.2.

$$\mathcal{L}_{rpn} = \mathcal{L}_{loc} + 0.2\mathcal{L}_{cont} \quad (7)$$

$$\mathcal{L}_{rpn} = \sum_t \sum_p L_{1_{smoot}} \left(\mathbf{d}_t^{p^*}, \mathbf{d}_t^p\right) + 0.2 \sum_t \sum_p (\mathbf{d}_t^p - \mathbf{d}_{t-1}^p) \quad (8)$$

The outputs of the $P \times D$ matrix are sent to the next deep learning module, LSTM.

### *G. Long Short-Term Memory Network (LSTM)*

Recurrent Neural Networks (RNN) [16] are a variant of the traditional neural network, which are highly effective at modeling temporal dependencies. However, RNN's are known to suffer from the vanishing/exploding gradient problem, which results in the loss of long-term dependencies [26]. Long Short-Term Memory (LSTM) solves this problem by providing key modifications. Namely, input, output, and forget gates are introduced to retain long term dependencies [27].

Because LSTM networks are so effective at modeling temporal relationships, they are optimal for many-to-many imaging; they can accumulate RF information over time to recover a *complete* pose. This is in comparison to direct 3D convolutions, which were shown by [36] to aggregate spatio-temporal information less effectively than 2D convolutions jointly trained with an LSTM. Therefore, the CNN spatial feature extractor and LSTM temporal architecture are explicitly separated in this architecture to encourage the learning of stronger spatio-temporal relationships.

At each timestep $t$, the LSTM outputs concatenated logit vectors $\mathbf{p}_t^p$, for all people $p$ in a frame. Then, each of the 15 keypoints $k$ on each person $p$ are transformed into a hidden representation for each keypoint $\mathbf{h}_{t_k}^p$ as:

$$\mathbf{h}_{t_k}^p = \mathbf{W}_k \mathbf{p}_t^p + \mathbf{b}_k, \ \forall k, \ p \quad (9)$$

Then, two sets of shared weights and biases are used to transform each hidden representation into a pixel classification: x-coordinate and y-coordinate. These weights and biases are shared across all keypoints; their sole responsibility is to transform an intermediate hidden representation into Cartesian coordinates.

$$\begin{cases} \alpha_{t_k}^p = \text{argmax}\left(\mathbf{W}_\alpha \mathbf{h}_{t_k}^p + \mathbf{b}_\alpha\right) \\ \beta_{t_k}^p = \text{argmax}\left(\mathbf{W}_\beta \mathbf{h}_{t_k}^p + \mathbf{b}_\beta\right) \end{cases} \forall k, p \quad (10)$$

where $\alpha$ represents the x-coordinate and $\beta$ the y-coordinate.

### *H. Context-Aware Region Proposals*

If multiple people are in a scene, accumulating information for each person requires tracking the identities of the people as time progresses. In [7], it was shown to be feasible to differentiate individuals based on their unique RF fingerprints. Therefore, we draw inspiration from the Siamese Neural Network [44] to teach the model to encode similar features for the same person in a sequence of RF images. In implementation, a new factor in the objective function is defined:

$$\mathcal{L}_{track} = \sum_t \sum_{p,q} \theta_{p,q} \left\| \boldsymbol{\sigma}_t^p - \boldsymbol{\sigma}_{t-1}^q \right\|_2 \quad (11)$$

where $\boldsymbol{\sigma}_b^a$ are the descriptive encoding features of the person $a$ at time $b$ and $\theta_{p,q}$ is an indicator value that is 1 if $p = q$ or $-1$ if not. In this manner, the minimization of $\mathcal{L}_{track}$ teaches the model to classify the same person to the same identity over time, and different people to different identities.

### *I. Novel Exponential Classification Objective*

With a high number of output pixels per timestep, overfitting is a significant risk. To combat this, a novel classification objective function $\mathcal{L}_{cls}$ is proposed by the researcher to encourage a better fit. $\mathcal{L}_{cls}$ is the direct sum between the individual losses for $\alpha$ and $\beta$, denoted as $\mathcal{L}_{cls_\alpha}$ and $\mathcal{L}_{cls_\beta}$, respectively. If $k$ is the keypoint being classified for person $p$ at time $t$, the two components of classification loss are given as:

$$
\begin{cases}
\mathcal{L}_{cls_\alpha} = \sum_t^T \sum_p^P \sum_k^K n_{t_k}^p e^{-\left(\log 2 - \delta_{t_k}^p\right)} \log p\left(\alpha_{t_k}^p \mid x_{t_k}^p[RF]\right) \\
\mathcal{L}_{cls_\beta} = \sum_t^T \sum_p^P \sum_k^K n_{t_k}^p e^{-\left(\log 2 - \delta_{t_k}^p\right)} \log p\left(\beta_{t_k}^p \mid x_{t_k}^p[RF]\right)
\end{cases} \tag{12}
$$

where $T$ is #timesteps, $P$ is #people, $K$ is #keypoints, $C$ is #classes. $\delta$ represents the L2 pixel distance between the predicted and ground-truth outputs:

$$
\delta_{t_k}^p = \sqrt{(\alpha_{t_k}^{p\,*} - \alpha_{t_k}^p)^2 + (\beta_{t_k}^{p\,*} - \beta_{t_k}^p)^2} \tag{13}
$$

$\mathcal{L}_{cls}$ is a customized Cross-Entropy classification loss [28] that adds an exponential term parameterized by $\delta_{t_k}^p$. It penalizes the model highly for misclassifying keypoints far from the ground truth and less so for closer mistakes, on an exponential scale. $\mathcal{L}$ also considers the confidence output of the cross-modal supervisory teacher model $n_{t_k}^p$ for each keypoint, making penalties more severe if the teacher label confidence is higher.

### J. Final Joint-Optimized Loss

We have now defined all 3 elements for the loss that is to be minimized: region proposal loss $\mathcal{L}_{rpn}$, tracking loss $\mathcal{L}_{track}$, and classification loss $\mathcal{L}_{cls}$ Through empirical experiments, the optimal loss balancing ratio between $\mathcal{L}_{rpn}$, $\mathcal{L}_{track}$, and $\mathcal{L}_{cls}$ was determined to be 4:3:4. Therefore, balancing coefficient $\lambda$ is established with value 0.75. Therefore, our final objective that jointly optimizes the weights of the CNN, RPN, and LSTM is:

$$
\mathcal{L} = \mathcal{L}_{rpn} + \lambda \mathcal{L}_{track} + \mathcal{L}_{cls} \tag{14}
$$

Adam [46] is adopted as the minimization algorithm on the compound objective defined in Equation 14.

### K. Researcher-Created Dataset With Adversarial Examples

75,000 training samples of {RF, RGB} images were collected at multiple locations including home, school, a fitness center, and a public library. Multiple mediums were also traversed: wood, brick, drywall, plastic, paper board, and open air were selected for testing. Data was partitioned 75/5/20 for training/validation/testing, respectively.

Table 2. Dataset Details

| | No Person | Walking | Gesturing | Sitting | Multi-Person | Total |
|---|---|---|---|---|---|---|
| Visible Scene | 3,750 | 18,750 | 11,250 | 3,750 | 7500 | 45000 |
| Occluded Scene | 3,750 | 11,250 | 7,500 | 3,750 | 3750 | 30,000 |
| Total | 7,500 | 30,000 | 18,750 | 7,500 | 11,250 | 75,000 |

The no-person scenes can be considered as adversarial examples, designed to fool the neural network. They they may contain RF signatures that could be mistaken for humans but are not. This alleviates the case of "phantom" predictions.

### L. Model Summary

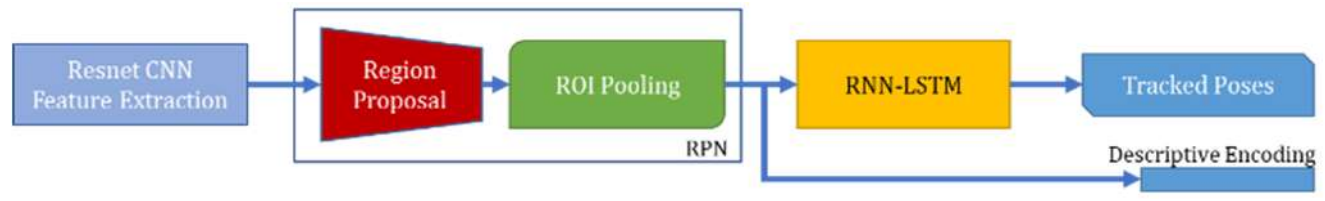

Figure 6. Summary

## IX. Output and Performance Analysis

### A. OKS Analysis

The Object Keypoint Similarity (OKS) [34] metric quantifies the average quality of keypoint localization over all 15 bodily keypoints defined in BODY-15. Keypoint Similarity (KS) is computed by evaluating an un-normalized Gaussian distribution. Where $\boldsymbol{\theta}^p$ are classified keypoints of person $p$, $\boldsymbol{\theta}^{p*}$ are ground truth locations, $C_i$ is the prediction's confidence level, and $s$ and $k_i$ are constants used to scale std. deviation:

$$
\begin{cases}
OKS(\boldsymbol{\theta}^{p*}, \boldsymbol{\theta}^p) = \dfrac{\sum_i KS\left(\boldsymbol{\theta}_i^{p*}, \boldsymbol{\theta}_i^p\right)\delta(C_i > 0)}{\sum_i \delta(C_i > 0)} \\
KS\left(\boldsymbol{\theta}_i^{p*}, \boldsymbol{\theta}_i^p\right) = e^{-\frac{\left\|\boldsymbol{\theta}_i^{p*} - \boldsymbol{\theta}_i^p\right\|_2^2}{2s^2k_i^2}}
\end{cases} \tag{15}
$$

Average Accuracy $AA_{ks}^p$ of classification for each person can be computed at various KS values: Given a threshold $T_{ks}$, compute the proportion of keypoints on person $p$ that fulfill $KS \geq T_{ks}$. $AA$ can also be computed per keypoint over the entire dataset. For context, ~95% of human-annotated keypoints will have $KS \geq .75$. As such, $KS = .75$ is considered a highly strict match; $KS = .50$ is a medium match.

#### i. Individual Keypoint Classification Error

General trends are analyzed for KS values of all keypoint types. The majority of keypoints are classified with low-to-none error. Only 6% of keypoints are classified with high error or missed classification, 30% hit medium error, and 64% achieved low-to-none error. Figure 7 shows a detailed breakdown. Shoulders and Hip are classified with highest consistency and accuracy, while hands and feet were most difficult to classify.

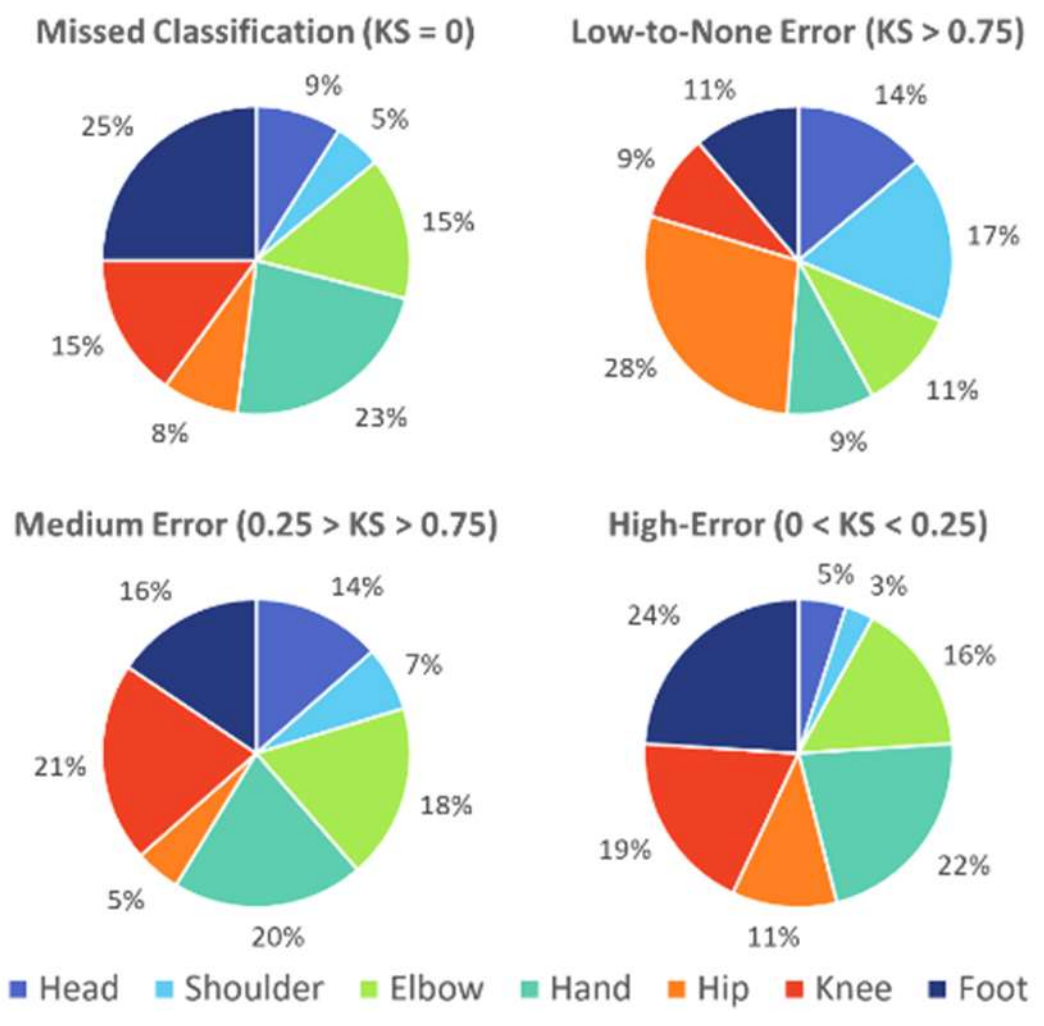

Figure 7. KS Performance by Keypoint

#### ii. Individual Keypoint AA on Various KS Levels & Obstruction Statuses

Putting the individual keypoint analyses together, AA for each keypoint is calculated at 2 differing threshold values: $KS_{50}$, and $KS_{75}$ (loose fit vs. strict fit). Data collected on both visible and through-obstruction scenes. Keypoints can be classified with high accuracy even at strict fit threshold.

Table 3. Keypoint AA at Various KS

| | Visible $KS_{50}$ | Obstructed $KS_{50}$ | Visible $KS_{75}$ | Obstructed $KS_{75}$ |
|---|---|---|---|---|
| **Head** | 99% | 91% | 70% | 50% |
| **Shoulder** | 99% | 95% | 86% | 72% |
| **Elbow** | 97% | 78% | 56% | 36% |
| **Hand** | 95% | 70% | 50% | 30% |
| **Hip** | 99% | 92% | 93% | 84% |
| **Knee** | 96% | 75% | 49% | 30% |
| **Foot** | 95% | 70% | 61% | 40% |

*iii. OKS Comparison with State-of-the-Art RGB Vision*

AA for RF pose parsing on visual scenes is compared to the reported values for Microsoft Human Pose Estimation (MSFT) [47], a state-of-the-art RGB image pose-decoding software. AA for RF pose parsing on through-obstruction scenes is measured.

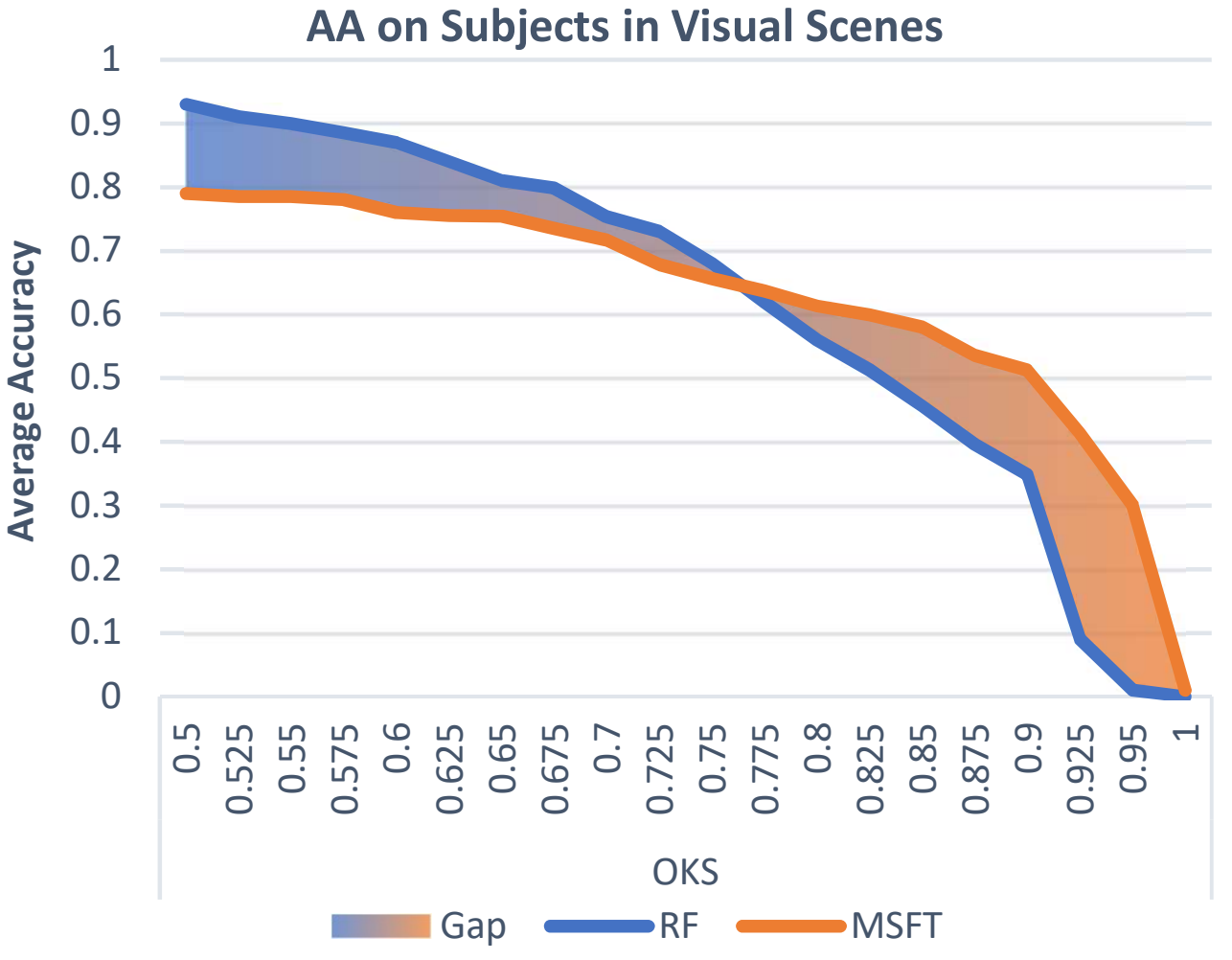

Figure 8. AA Over Multiple OKS

RF outperforms MSFT on low-threshold classification. Possible reasons include that the novel objective function (Equation 13) reduces overfitting, and that we employ many-to-many imaging using LSTM RNN's (§VIII-G) to consider multiple frames of input; this contrasts with the frame-by-frame processing approach for MSFT.

RF has high performance even for through-obstruction inference. Possible explanations include the novel training pipeline introduced in §VIII-C and the many-to-many imaging technique detailed in §VIII-G.

*iv. Many-to-Many Analysis*

The performance improvement attained when applying the new Many-to-Many imaging methodology is evaluated in this section. The See-Through Model is repurposed to process RF images frame-by-frame without consideration to previous context. Concretely, the LSTM model is removed while the CNN and RPN operate on a frame-by-frame basis, sending their features to fully connected layers fine-tuned to output pose predictions for each individual frame. Many-to-Many Imaging provides a key improvement over a frame-by-frame approach.

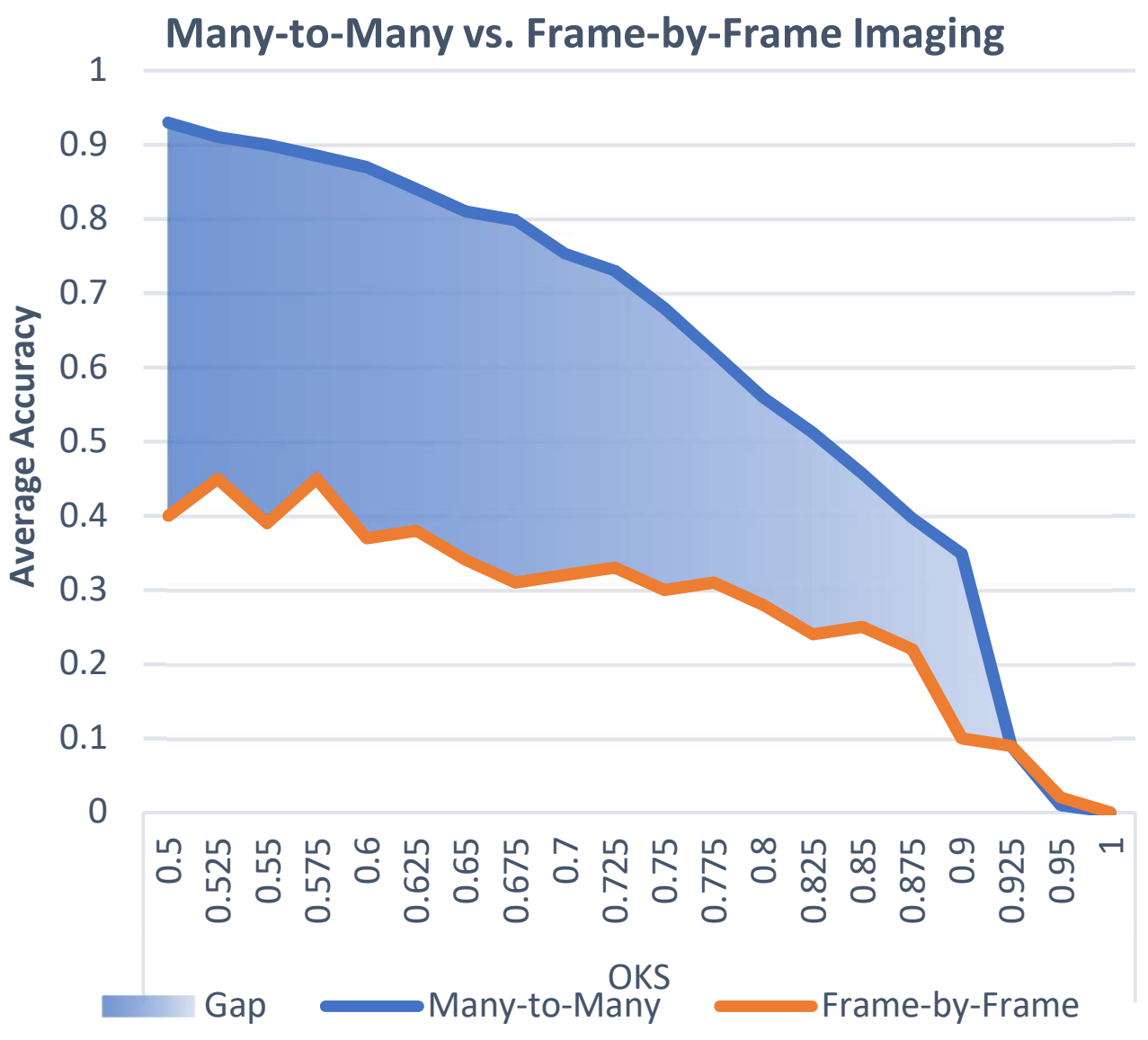

Figure 9. Many-to-Many Analysis

*v. Medium Analysis*

The following graph compares AA through different types of obstruction. Typical types of obstruction have minimal impact on performance depending on their dielectric constant.

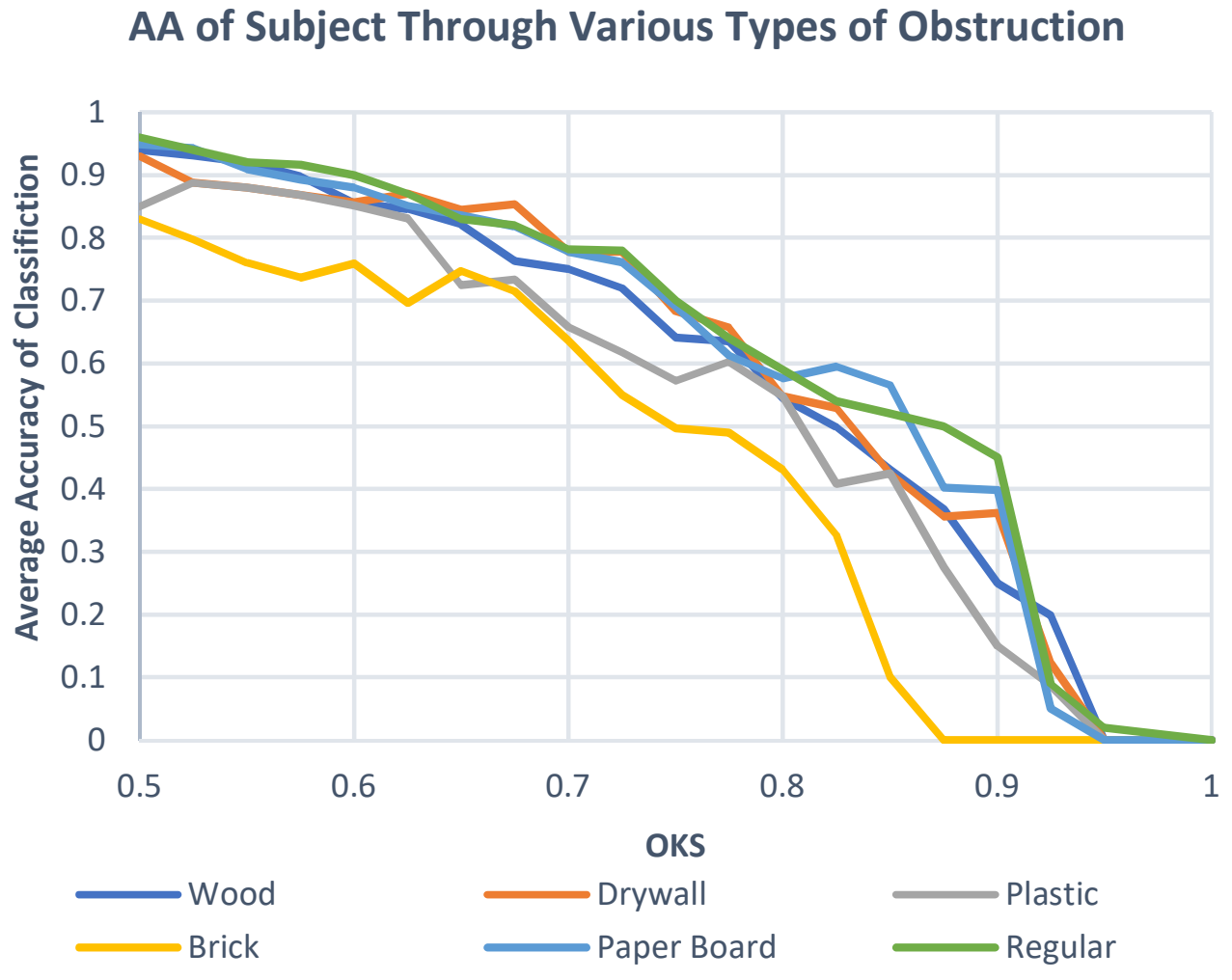

Figure 10. Medium Analysis

*vi. Real-Time Setup Analysis*

System execution was timed to verify the real-time capability of the system. For 100,000 frames to be processed, 5,565.153 seconds were taken. This amounts to 0.056 sec per frame, which is sufficiently fast to power a 5Hz stream of data. It also leaves roughly 0.15 sec for network latency in the field.

## X. Conclusion

The key contribution of this study is the proposal of a many-to-many RF imaging methodology, in contrast to one-to-one imaging for optical systems, to extend the boundaries of human vision beyond visual obstructions. This overcomes the challenge of specular reflection by exploiting movement and synthesizing time-sequential data. One particularly notable application of this work is the detection of people trapped in burning buildings or foliage of the forest from safety.

Contributions are summarized as follows: 1) Created a many-to-many imaging decoder using state-of-the-art deep learning techniques. 2) Proposed a novel objective function for anti-overfitting optimization. 3) Implemented a new training pipeline to solve a bottleneck with inexplicit through-wall training.

## XI. Demo Simulator & Video

A demonstration video was filmed to show both offline and online processing: `https://youtu.be/7hX8qGJdWno`. A screenshot can be seen below:

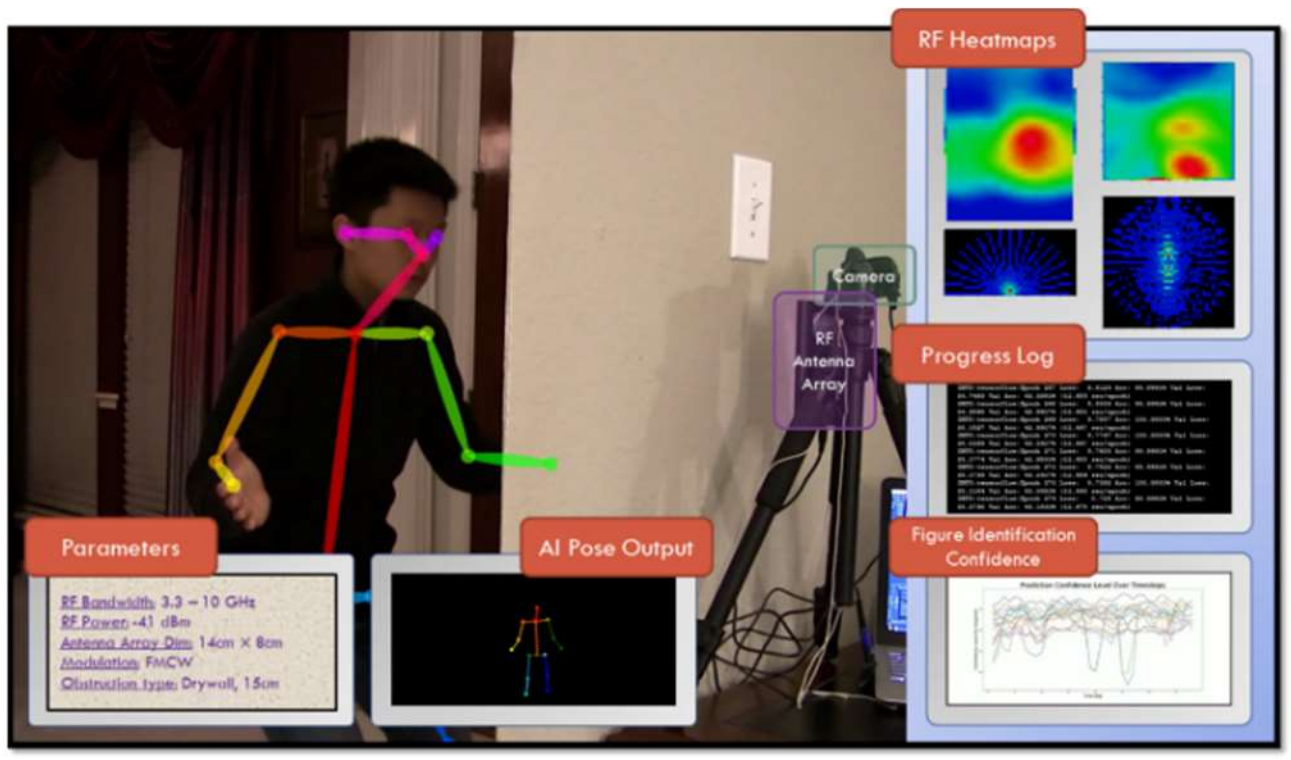

## XII. Future Research

In the future, the principles established in this paper can be augmented using multiple RF Antenna Arrays to achieve stereoscopic RF Imaging. RF transceivers can be arranged in an arc to capture a larger subset of reflections. This could possibly mitigate the Body-Part Specularity challenge and allow for a more accurate and complete pose construction.